\documentclass[twoside]{article}
\usepackage[accepted]{aistats2018}
\usepackage{amsmath,amssymb,amsfonts,amsthm,latexsym}
\usepackage{graphicx,xcolor}
\usepackage{ifthen}

\newcommand{\Reals}{\mathbb R}

\newcommand{\Integers}{\mathbb Z}
\newcommand{\IntegersP}{\Integers_+}
\newcommand{\Prob}{\mathbb{P}}
\newcommand{\E}{\mathbb{E}}

\newcommand{\argmax}{\operatorname{argmax}}
\newcommand{\ba}{\mathbf{a}}

\newcommand{\cE}{\mathcal{E}}
\newcommand{\cF}{\mathcal{F}}

\newcommand{\cI}{\mathcal{I}}
\newcommand{\cN}{\mathcal{N}}

\newcommand{\cQ}{\mathcal{Q}}
\newcommand{\cS}{\mathcal{S}}

\newcommand{\cV}{\mathcal{V}}
\newcommand{\cW}{\mathcal{W}}

\newcommand{\Ind}{\mathbb{I}}

\newcommand{\MSE}{\text{\normalfont MSE}}
\newcommand{\sgn}{\text{\normalfont sign}}

\theoremstyle{plain}
\newtheorem{allcnt}{AllCnt}[section]

\newtheorem{theorem}[allcnt]{Theorem}

\theoremstyle{definition}

\newtheorem{model}[allcnt]{Model}

\theoremstyle{remark}

\newboolean{showcomments}
\setboolean{showcomments}{false}
\newcommand{\comment}[1]{  \ifthenelse{\boolean{showcomments}}
{ \textcolor{red}{#1}} {}  }

\newboolean{showproofs}
\setboolean{showproofs}{true}
\newcommand{\pf}[1]{\ifthenelse{\boolean{showproofs}}
{#1}{}}

\begin{document}

%

%

\twocolumn[
\aistatstitle{Reducing Crowdsourcing to Graphon Estimation, Statistically}

\aistatsauthor{ Devavrat Shah \And Christina Lee Yu}
\aistatsaddress{ Massachusetts Institute of Technology \And  Microsoft Research New England } 
]

\begin{abstract}
Inferring the correct answers to binary tasks based on multiple noisy answers in an unsupervised manner has emerged as the canonical question for micro-task crowdsourcing or more generally aggregating opinions. In graphon estimation, one is interested in estimating edge intensities or probabilities between nodes using a single snapshot of a graph realization. In the recent literature, there has been exciting development within both of these topics. In the context of crowdsourcing, the key intellectual challenge is to understand whether a given task can be more accurately denoised by aggregating answers collected from other different tasks. In the context of graphon estimation, precise information limits and estimation algorithms remain of interest.  

In this paper, we utilize a statistical reduction from crowdsourcing to graphon estimation to advance the state-of-art for both of these challenges. We use concepts from graphon estimation to design an algorithm that achieves better performance than the {\em majority voting} scheme for a setup that goes beyond the {\em rank one} models considered in the literature. We use known lower bounds for crowdsourcing to derive lower bounds for graphon estimation. 
\end{abstract}

\section{Introduction}

\noindent{\bf Crowdsourcing: background.}
In recent years, crowd-sourcing platforms have become complementary computing systems to scale tasks that are difficult for algorithms to solve, but easy and trivial for humans. For example, this includes tasks such as image recognition (``is this picture culturally acceptable''), content censorship (``is this webpage suitable for children'') or social opinion (``is this a good coffee shop for writing a paper''). It may be computationally challenging to train an algorithm to determine if an image is culturally offensive, or if a webpage contains explicit content; and it would be impossible for an algorithm to provide a human opinion on the suitability of a coffee shop for writing a paper, or on the pros and cons of legalizing marijuana. However, a human could relatively easily and quickly provide answers to these  tasks. 

As a result, crowd-sourcing platforms such as Amazon Mechanical Turk have emerged, on which requesters post tasks that they would like to be solved along with a monetary reward for completion, and human workers browse the posted tasks and earn money for providing responses to these tasks. For a variety of reasons, the responses provided by human workers may not be consistent amongst themselves, and may not correspond to the true answer or solution for the task. For example, workers may have different levels of language proficiency, leading to noisy responses for language translation tasks. Alternatively, even if a worker is capable of solving the task, s/he may be lazy and may respond arbitrarily to save effort. In the context of collecting social opinion, lack of consensus in the population responding to the task or question is expected. Therefore, the challenge is, given a set of responses provided by human workers for a set of tasks that are noisy, unreliable and potentially contradicting each other, can we infer the true answer or solution for the task? 

\footnotetext{Author names appear in alphabetical order of their last names.}

\noindent{\bf Crowdsourcing: formally.} Let there be $T \geq 1$ binary tasks with $a_i \in \{-1, 1\}$ for $i \in [T]$ representing the {\em true} answer for the $i$th task\footnote{We shall use notation $[N] \equiv \{1,\dots, N\}$.}. Let there be $W \geq 1$ workers. If worker $j $ is asked to provide answer to task $i$, then it will be denoted $M_{ij}$, which has distribution 
\begin{align}
M_{ij} & = \begin{cases} 
a_i &\mbox{with probability } F_{ij} \\ 
-a_i &\mbox{with probability } 1 - F_{ij}, 
	\end{cases} 
	\label{eq:model}
\end{align}
where $F_{ij} \in [0,1]$ represents the ``skill'' of worker $j$ for task $i$. Therefore, $\E[M_{ij}] = a_i (2 F_{ij} - 1)$. We shall assume that $M_{ij}$ are independent across all $i,j$. We shall use notation $M = [M_{ij}] \in \{-1,1\}^{T \times W}$, $\ba = [a_i] \in \{-1,1\}^{T}$ and $F = [F_{ij}] \in [0,1]^{T \times W}$.

As a crowdsourcing system operator, our decision is two fold: (a) determine a minimal assignment of tasks to workers denoted by the subset $\cE \subset [T] \times[W]$ where $M_{ij}$ is then queried for all $(i,j) \in \cE$, and (b) determine an algorithm to infer $a_i, ~i\in [T]$ so that the fraction of answers we infer incorrectly is minimized. Naturally, the design for (a) and (b) will collectively determine the trade-off of between the accuracy of inference and number of queries per task. The eventual goal is to achieve the {\em pareto} boundary of this trade-off. 

\noindent{\bf Crowdsourcing: prior work.} We describe the prior work that have attempted to understand the above trade-off between accuracy and the number of worker queries per task by imposing structure on $F$ (and hence $\E[M]$). The first formal model (an instance of \eqref{eq:model}) was proposed by Dawid and Skene \cite{DawidSkene79} where $F_{ij} = w_j$ for all $i \in [T]$, with $w_j \in [0,1]$ representing the ``skill'' of worker $j \in [W]$. Various inference algorithms were proposed including EM algorithm \cite{DawidSkene79, GaoZhou13, ZhangChenZhouJordan14}, belief propagation and iterative methods \cite{KargerOhShah11, KargerOhShah13, KargerOhShah14, LiuPengIhler12, LiYu14, OkOhShunYi16}, and spectral methods \cite{GhoshKaleMcAfee11, DalviDasguptaKumarRastogi13,BonaldCombes16}. The minimax optimal trade-off curve for this model states that to achieve accuracy $\alpha$, the number of worker queries per task requires a scaling of $(\frac{1}{W}\sum_j (2w_j - 1)^2)^{-1} \ln(\frac{1}{\alpha}))$ \cite{KargerOhShah11, KargerOhShah13, KargerOhShah14}, which does not change even when one has the option to query adaptively.  

The simple {\em majority voting} algorithm where the we infer the answer to a given task as simply the majority of the answers provided to it, provides a trade-off such that to achieve accuracy $\alpha$, the number of worker queries per task requires a scaling of $(\frac{1}{W}\sum_j (2w_j - 1))^{-2} \ln(\frac{1}{\alpha}))$. Therefore, if the worker skills are not identical, then majority vote strictly requires more queries compared to the optimal algorithm. Put it other way, there is a {\em strict benefit} in utilizing answers of workers for different tasks to better estimate answers for a given task for the Dawid-Skene model. Effectively, this has led to the following intellectual quest: 

{\em Under what conditions on $F$ (or $\E[M]$) can combining answers of workers for all tasks help, i.e. when is it possible to beat majority voting?}

Surprisingly, this has turned out to be a hard question. The notable progress towards this quest have been in recent works \cite{KhetanOh16, ShahBalakrishnanWainwright16} where authors allow the $F_{ij}$ to depend on the task difficulty in addition to worker skill as in Dawid-Skene model. They show that it is indeed possible to beat majority voting in specific settings; however, all of the models for which an algorithm has been formally established to beat majority vote, result in assumptions that limit $\E[M]$ to a {\em rank one} matrix. We ask the question: {\em is it possible to beat majority voting when $\E[M]$ has rank $d > 1$?}

\noindent{\bf Crowdsourcing: our contribution.} We answer the above question in the affirmative. We show that for a natural model where the workers and tasks are of $d$ types, with $F_{ij} = p$ if $i$ and $j$ are same type, and $F_{ij} = \tfrac{1}{2}$ otherwise, then a simple algorithm achieves accuracy $\alpha$ with the number of worker queries per task scaling as $(\frac{d}{(2p-1)^{2}} \ln(\frac{d}{\alpha}))$; in contrast, majority voting requires $(\frac{d^2}{(2p-1)^{2}} \ln(\frac{1}{\alpha}))$ queries per task. 

\noindent{\bf Graphon: formally, prior work.} In graphon or $1$-bit matrix estimation, we partially observe a symmetric matrix $Y = [Y_{ij}] \in \{0,1\}^{n \times n}$. Specifically, we observe entries of $Y$ for $\cE \subset [n] \times [n]$, i.e. for all $(i,j) \in \cE$, we observe $Y_{ij}$. The matrix $Y$ is generated so that $Y_{ij}$ are independent. The goal is to recover the expected matrix $\E[Y] \in [0,1]^{n \times n}$ by observing as few entries ($= |\cE|$) of $Y$ as possible. In the setting of graphon estimation, one assumes that each $i \in [n]$ is associated with a {\em latent} parameter $\theta_i \sim U(0,1)$ sampled uniformly on the unit interval independently across all $i$. Then $\E[Y_{ij}] = f(\theta_i, \theta_j)$ for some measurable {\em latent} function $f: [0,1]\times [0,1]\to [0,1]$ for all $i, j \in [n]$. The fraction of observations, or samples from the ${n \choose 2}$ unique points in matrix $Y$, that is required to recover $P$ accurately depends on the structure of $f$. 

There has been lots of recent progress in answering this question for different structures of $f$. For the stochastic block model where $\E[Y]$ has finitely many distinct rows (and columns), very precise understanding has been obtained, cf. \cite{AbbeSandon15a, AbbeSandon15b, ChinRaoVu15}. For $P$ with strictly monotonic row and column sums, \cite{ChanAiroldi14, YangHanAiroldi14, BorgsChayesCohnGanguly15} provide estimators that are consistent with minimal observations required. If $f$ is Lipschitz, there are consistent estimators for both graphon estimation and matrix completion (cf. \cite{ZhangLevinaZhu15, Chatterjee15, song2016blind, BorgsChayesLeeShah17}). Results in matrix completion and mixed membership stochastic block model estimation provide estimators when $\E[Y]$ is low rank (cf. \cite{Chatterjee15,ChenWainwright15,DavenportPlanBergWootters14,AnandkumarGeHsuKakade13} ). For the most general setting in which the latent function $f$ can be any measurable function, there do not yet exist polynomial time estimators, although \cite{KloppTsybakovVerzelen15} has studied the rate achieved by a least squares estimator, and provided accompanying minimax convergence rates.

Despite this remarkable progress, lower bounds for graphon estimation are still being developed. In the recent work of \cite{SteurerHopkins17}, for the mixed membership model, an explicit lower bound has been conjectured for the detection of communities.  Establishing such lower bounds unconditionally can lead to resolution of statistical and computational trade-offs. \cite{Xu17} and \cite{KloppTsybakovVerzelen15} have shown that the minimax optimal rate for graphon estimation with $\alpha$-Holder smoothness is 1 for $np = O(1)$, $\log(np)/np$ for $\log(np) \leq \alpha \log n + (\alpha+1)\log\log n$, and $(n^2 p)^{-\alpha/(\alpha+1)}$ for $\log(np) \geq \alpha \log n + (\alpha+1)\log\log n$.

\noindent{\bf Graphon: our contribution.} By providing an explicit statistical reduction of crowdsourcing to graphon estimation, we transfer a lower bound utilized in \cite{KargerOhShah11, KargerOhShah13, KargerOhShah14} to obtain lower bounds for graphon estimation. In \cite{KargerOhShah11, KargerOhShah13, KargerOhShah14}, this lower bound was crucial for establishing the minimax optimality for the Dawid-Skene model. The implied lower bound suggests an invariant that the {\em number of observations multiplied by the square of the minimal eigenvalue of $f$ with respect to its spectral decomposition must be larger than a universal constant}. This lower bound provides a analytic form for an ``incoherence'' like assumption that has been made popular in the matrix estimation literature. In comparison to the previous lower bound, our bound is looser as a function of $n$ and $p$, but it shows the dependence on spectral properties such as the smallest eigenvalue and eigenfunction amplitude, which is omitted in the previous bounds.


We present our contribution in the context of graphon parameter estimation because it is a recently active area of particular interest which emphasizes the binary data model. However our results relate to the broader setting of asymmetric matrix estimation as well.

\section{Metrics, Models and Baselines}

\subsection{Metrics}

\noindent{\bf Crowdsourcing.} We shall measure the performance of an inference algorithm in terms of the fraction of answers obtained correctly. Precisely, given true answers $\ba = [a_i] \in \{-1,1\}^T$, the error of inferred answers $\hat{\ba} = [\hat{a}_i]\in \{-1,1\}^T$ is 
\begin{align}
\mbox{error}(\hat{\ba}, \ba) & = \tfrac{1}{T} \Big(\textstyle\sum_{i=1}^T \Ind(\hat{a}_i \neq a_i)\Big),
\end{align}
where $\Ind(\cdot)$ is the indicator function. The number of queries per task is simply $|\cE| / T$. The goal is to determine
what is the minimal number of queries per task needed to obtain $\E[\mbox{error}(\hat{\ba}, \ba)] \leq \alpha$ for any $\alpha \in (0,1)$.

\noindent{\bf Graphon.} The accuracy of an estimate $\hat{Y}$ for the matrix $\E[Y]$ is measured in terms of the Mean-Squared-Error (MSE) defined as
\begin{align}
\MSE & = \E\Big[\tfrac{1}{n^2} \textstyle\sum_{i,j} (\hat{Y}_{ij} - \E[Y_{ij}])^2 \Big].
\end{align}
We would like to produce an estimate $\hat{Y}$ given observed entries $\{Y_{ij}\}_{(i,j) \in \cE}$, indexed by $\cE \subset [n]\times [n]$, which denotes the observed locations of matrix $Y = [Y_{ij}]$. We assume the entries are uniformly sampled, i.e. $\cE$ includes each of the possible entries with probability $p$, independently, such that on average $|\cE| = n^2 p$. The question is, what is the minimal sample probability $p$ that is required to achieve $\MSE \leq \alpha$. 

\subsection{Models}

\noindent{\bf Prior Models Are Rank $1$.} There are three prior works that provide theoretical analyses establishing that the performance of the {\em majority estimation algorithm can be beaten} in the context of crowdsourcing. We describe those three models below. As we shall see, for each of these three models, the expected response matrix $\E[M]$ turns out to be rank $1$.

\begin{model} [Dawid-Skene] This model assumes every task is homogeneous;  and worker $j$ is associated with a reliability parameter $w_j \in [0,1]$ for $j \in [W]$, which is the probability a worker solves the task correctly \cite{DawidSkene79}. The probability that worker $j$ answers task $i$ correctly is given by $F_{ij} = w_j$. Therefore, $\E[M_{ij}] = a_i (2w_j - 1)$, such that $\E[M]$ is a rank $1$ matrix.
\end{model}

\begin{model}
In \cite{KhetanOh16}, every task $i$ is associated with a difficulty parameter $t_i \in \left[\tfrac{1}{2},1\right]$, i.e. the probability a task is perceived incorrectly, and every worker $j$ is associated with a reliability parameter $w_j \in [0,1]$, i.e. the probability a worker responds consistently with perceived answer. The probability worker $j$ answers task $i$ correctly is given by $F_{ij} = t_i w_j + (1-t_i)(1 - w_j)$. Therefore, $\E[M]$ is a rank $1$ matrix, 
\begin{align*} 
\E[M_{ij}] & = a_i (2F_{ij} - 1) ~=~(a_i(2t_i-1)) \times (2w_j-1).
\end{align*}
\end{model}

\begin{model}
In \cite{ShahBalakrishnanWainwright16},  it is assumed that there is an ordering (or permutation) of worker skill levels and task difficulty levels under which the probability of a correct response is monotonic. Specifically, their theoretical guarantees hold for the model $F_{ij} = w_j (1 - t_i) + \tfrac12 t_i$, which also implies $\E[M]$ is a rank 1 matrix,
\begin{align*} 
\E[M_{ij}] & = a_i (2F_{ij} - 1) ~=~(a_i(1-t_i)) \times (2w_j - 1).
\end{align*}
\end{model}

There are other models considered in the literature such as \cite{HoVaughan12, HoJabbariVaughan13} assuming that each task (and hence worker) are of $d$ different types. It does not result in $\E[M]$ being rank $1$, but their theoretical analysis assumes the knowledge of the correct answers for a subset of tasks in order to bootstrap the answer. In \cite{WelinderBransonBelongiePerona10}, the $F_{ij}$ is derived from a Gaussian distribution. In effect, $F_{ij} = f(t_i, w_j)$ where $f$ is a Lipschitz function. They do not provide a theoretical analysis for how their approach compares to majority voting. \cite{SteinhardtValiantCharikar16} considers a model in which there are three types of collected responses: ratings provided by a large fraction of reliable workers, ratings from bounded fraction of adversarial workers, and stochastic ratings obtained ``ourselves''. Their model allows for adversarial responses which is beyond a purely stochastic model. Their reliable workers provide answers which are monotonically increasing with the quality, similar to the monotonic model of \cite{ShahBalakrishnanWainwright16}. However their model still does not allow for specialization of different types of workers for tasks, where workers may be good or bad depending on the type of task.

\noindent{\bf Rank $d$ Model.} In this work, we shall argue that it is possible to beat majority voting even when $\E[M]$ has rank $d$ with $d > 1$. We use a simple model that leads to a rank $d$ structure of $\E[M]$. It is similar to the classical stochastic block model with $d$ communities.

\begin{model}[$d$-Type Specialization Model]
Each task $i \in [T]$ is associated with a task type $t_i \in [d]$, each worker $j \in [W]$ is associated with a worker type $w_j \in [d]$. Assume the probability that worker $j$ correctly answers task $i$ is denoted $F_{ij}$, which takes value $p$ if $t_i = w_j$, and $\tfrac12$ otherwise. It follows that $\E[M_{ij}] = 2p - \tfrac{1}{2}$ if $t_i = w_j$ and $\E[M_{ij}] = 0$ otherwise.
$\E[M]$ is equivalent to a rank $d$ block matrix with $d$ disjoint blocks. We assume worker types are sampled uniformly amongst the $[d]$ types and task types are uniformly drawn from the $d$ types. Let $\{\cW_1, \dots \cW_d\}$ partition the workers into their associated types such that $w_j = z$ for all $j \in \cW_z$. We assume $T \gg W$. 
\end{model}

\subsection{Baselines for Performance}

We discuss the majority voting estimation algorithm and its performance. This serves as the baseline for any crowd-sourcing algorithm since majority voting only utilizes the answers associated with a given task to make inference. We also discuss the improvement that could be gained by a maximum likelihood oracle with extra knowledge of the parameter matrix $F$.

\noindent{\bf Majority voting.}
Let $\cN(i) = \{j: (i,j) \in \cE\}$ denote the set of workers which are assigned to task $i \in [T]$. Define the majority vote estimate for $a_i$ as 
\begin{align}
\hat{a}^{MV}_i = \sgn\left(\textstyle\sum_{j \in \cN(i)} M_{ij}\right).
\end{align}
Under either of the assumptions that $F_{ij} > \tfrac12$ for all $(i,j)$ or simply that $\textstyle\sum_{j \in \cN(i)} F_{ij} > \tfrac12$ for all $i$, a simple
application of Chernoff's bound implies 
\begin{align*}
\Prob(\hat{a}^{MV}_i \neq a_i) & \leq \exp\Big(-\tfrac{|\cN(i)|}{2}\Big(\tfrac{\textstyle\sum_{j \in \cN(i)} (2 F_{ij} - 1)}{|\cN(i)|}\Big)^2\Big).
\end{align*}
If we choose $\cN(i) \subset [W]$ at random, effectively 
\begin{align*}
\Prob(\hat{a}^{MV}_i \neq a_i) & \leq \exp\Big(-\tfrac{|\cN(i)|}{2} (2\E[F_{i \star}] - 1)^2\Big),
\end{align*}
where $\E[F_{i \star}]$ denotes the expected value of $F_{ij}$ assuming that worker $j$ is chosen uniformly at random amongst $[W]$.
To achieve $\Prob(\hat{a}^{MV}_i \neq a_i)  \leq \alpha$, we assign
$|\cN(i)|  \approx \tfrac{2}{(2\E[F_{i \star}] - 1)^2} \ln(\tfrac{1}{\alpha})$
workers per task. For the $d$-Type Specialization Model, this requires
\begin{align}\label{eq:majvote}
|\cN(i)| & = \Theta\Big(\tfrac{d^2}{(2p-1)^2} \ln\left(\tfrac{1}{\alpha}\right)\Big).
\end{align}

\noindent{\bf Maximum Likelihood Oracle.}
Suppose we had an oracle that told us the parameter matrix $F$. How much can we improve the estimator? We can simply compute the maximum likelihood solution,
\begin{align}
\hat{a}^{ML}_i = \sgn\left(\textstyle\sum_{j \in \cN(i)} \ln\left(\tfrac{F_{ij}}{1 - F_{ij}}\right) M_{ij}\right).
\end{align}
Let us denote $\omega_{ij} = \ln(\tfrac{F_{ij}}{1 - F_{ij}})$. We can verify that $\sgn(\E[\omega_{ij} M_{ij}]) = a_i$, and the magnitude increases as $F_{ij}$ is more polarized towards 0 or 1. If $F_{ij} = 1$, then $\omega_{ij} = \infty$, and if $F_{ij} = 0$, then $\omega_{ij} = -\infty$, such that a single datapoint would dominate the estimate. In either of these two cases, the probability of error is zero, since at least one worker is fully reliable. Otherwise, the probability of error is bounded by
\begin{align*}
\Prob\Big( \Big|\textstyle\sum_{j \in \cN(i)} \omega_{ij} (M_{ij} - \E[M_{ij}]) \Big| \geq \textstyle\sum_{j \in \cN(i)} \omega_{ij} \E[M_{ij}] \Big).
\end{align*}
Clearly, even if we had $F_{ij} =1$ for even one worker $j$ for a given task $i$, then maximum likelihood could outperform majority voting drastically. But issue is that
we {\em do not} know $F$. So the question remains: {\em when can we beat majority vote?} 

\noindent{\bf Directly Using Graphon Estimation.} We can approximate the maximum likelihood oracle by estimating the matrix $\E[M]$ directly from data using methods from graphon estimation, and then using it to estimate $\ba$. Let $\hat{M}$ be the estimate of $\E[M]$ obtained from the observed data matrix $M$. Under the assumption that $\sum_{j \in [W]} F_{ij} > \tfrac12$, this leads to estimates of
\begin{align}
\hat{a}_i & = \sgn(\textstyle\sum_{j \in [W]} \hat{M}_{ij}) \quad i \in [T].
\end{align}
Hence, $\Prob(\hat{a}_i \neq a_i)$ is bounded above by
\[\Prob\Big( \Big|\textstyle\sum_{j \in [W]} (\hat{M}_{ij} - \E[M_{ij}])\Big| \geq \textstyle\sum_{j \in [W]} \E[M_{ij}] \Big). \]
Observe that we can even include inferred values $\hat{M}_{ij}$ for $(i,j) \notin \cE$, for which the response $M_{ij}$ was not even observed. Evaluating this error bound expression will depend on the spectral properties of the model and the matrix estimation algorithm. 

Under the assumption that $F_{ij} > \tfrac12$, an estimation algorithm could be 
\begin{align}
\hat{a}_i = \sgn(\hat{M}_{ij^*}) \text{  for  } j^* \in \argmax_j |\hat{M}_{ij}|. \label{eq:max_est}
\end{align}
If $|\hat{M}_{ij} - \E[M_{ij}]| \leq \tfrac{1}{2}(\max_j |\E[M_{ij}]| + \min_j |\E[M_{ij}]|)$ for all $(i,j)$, then $\hat{a}_i = a_i$. Therefore $\Prob(\hat{a}_i \neq a_i)$ is bounded by
\begin{align*} 
&\Prob\Big(\max_{j \in [W]} |\hat{M}_{ij} - \E[M_{ij}]| \geq \tfrac{1}{2}\Big(\max_j |\E[M_{ij}]| + \min_j |\E[M_{ij}]|\Big) \Big). 
\end{align*}
This evaluation depends on the graphon estimation algorithm (and $\E[M]$), which unfortunately is more difficult. Effectively, crowdsourcing requires estimating only the {\em sign} of a column sum of $\E[M]$ while graphon estimation requires estimating the entire matrix. As a result, the naive use of a graphon estimator is unlikely to beat majority vote. This is precisely the challenge we overcome in this paper. 

\section{Result I: Beating Majority}

We describe how we can beat majority voting for the $d$-Type Specialization Model using a simple graphon estimation algorithm coupled with a clever crowdsourcing design (assignment of workers to tasks). We state the theorem below, followed by a description of the algorithm that achieves the theorem and the corresponding proof. 

\begin{theorem}\label{thm:main.1}
Assuming the $d$-Type Specialization model, for any given $\alpha \in (0,1)$ and $T \gg W$, 
there exists a design and estimation algorithm for crowdsourcing such that by soliciting no more than 
\[\frac{16 d}{(2p - 1)^2} \ln\left(\frac{6 d}{\alpha}\right) \]
responses per task, the  estimation $\hat{\ba}$ of $\ba$ is such that 
\begin{align*}
\E[\mbox{\em error}(\hat{\ba}, \ba)] \leq \alpha.
\end{align*}
\end{theorem}

\noindent{\bf Two Stage Algorithm.} We describe the algorithm that leads to Theorem \ref{thm:main.1} for the $d$-Type Specialization model. The algorithm takes parameter $R, L$ and $\xi$ that will be determined
later. 
\begin{itemize}
\item Stage 1: Clustering Workers by Types.
\begin{itemize}
\item[$\circ$] Let $\cS \subset [T]$ represent randomly chosen $R$ tasks from $T$ tasks, i.e. $|\cS| = R$. 
\item[$\circ$] Assign {\em all} $W$ workers to solve tasks in $\cS$. 
\item[$\circ$] For $j \in [W]$, assign them to (disjoint) clusters of workers sequentially as follows: 
\begin{itemize}
\item[$\bullet$] If there exists a cluster of workers $\cQ \subset [j-1]$ such that for each $j' \in \cQ$
\[\tfrac{1}{R} \textstyle\sum_{i \in \cS} \Ind(M_{ij} = M_{ij'}) > \xi,\]
then assign $j$ to (any) such cluster $\cQ$. 
\item[$\bullet$] If no such cluster $\cQ$ exists, create a new cluster containing $j$, i.e. $\{j\}$.
\end{itemize}
\item[$\circ$] Let $C$ be the number of disjoint clusters constructed in the previous step, and let $\{\cV_1 \dots \cV_C\}$ denote the constructed clusters, which forms a disjoint partitioning of $[W]$.
\item[$\circ$] Given the above estimated worker clusters, for each task $i \in [T] \setminus S$ and for each cluster $z \in [C]$, 
assign task $i$ to $L$ workers sampled uniformly at random amongst the set of workers $\cV_z$. Therefore, each task $i$ is assigned to 
a total of $LC$ workers.
\end{itemize}
\item Stage 2: Estimating the Task Answers
\begin{itemize}
\item[$\circ$] Let $\cN(i) \subset [W]$ denote the workers assigned to task $i$. 
Let 
$z^*(i) = \argmax_{z \in [C]} |\textstyle\sum_{j \in \cN(i) \cap \cV_z} M_{ij} |.$
\item[$\circ$] Then
$\hat{a}_i = \sgn\left(\textstyle\sum_{j \in \cN(i) \cap \cV_{z^*(i)}} M_{ij}\right).$
\end{itemize}
\end{itemize}

\noindent{\bf Proof of Theorem \ref{thm:main.1}.} We choose the
following parameters for the proof. Let
$\xi = \tfrac12 + \tfrac{( 2 p - 1)^2}{4 d},
R = \tfrac{8 d^2}{ ( 2 p - 1)^4} \ln\left(\tfrac{3W(W-1)}{2\alpha}\right),
L = \tfrac{8}{(2p - 1)^2} \ln\left(\tfrac{6 d}{\alpha}\right),
W \geq \tfrac{16d}{(2p - 1)^2} \ln\left(\tfrac{6 d}{\alpha}\right),$ 
and $T \geq C' W^3,$
where $C'$ is a large enough constant so that with the above choices of parameters (esp. $W$),
we have that 
\begin{align*}
\tfrac{8 W d^2}{ T ( 2 p - 1)^4} \ln\left(\tfrac{3W(W-1)}{2\alpha}\right) & \leq 
\tfrac{8d}{(2p - 1)^2} \ln\left(\tfrac{6d}{\alpha}\right).
\end{align*}
First we analyze the result of partitioning the workers into clusters. For a pair of workers $(a,b)$, the expected fraction of matching responses will be larger if the workers have the same type as opposed to different types. Consider a pair of workers $(a,b)$ whose type is equal. For each assigned task $i \in S$, with probability $1/d$, the task type is equal to the worker type, such that the probability that $M_{ia} = M_{ib}$ is equal to $p^2 + (1 - p)^2$. With probability $1 - 1/d$, the task type is different from the worker type, such that the probability that $M_{ia} = M_{ib}$ is equal to $\tfrac12$. Therefore, the indicator random variable $\Ind(M_{ia} = M_{ib})$ is a Bernoulli random variable with parameter 
\[\tfrac{p^2 + (1 - p)^2}{d} + \tfrac{(d-1)}{2d} = \tfrac12 + \tfrac{(2 p - 1)^2}{2d}.\]
If the pair of workers $(a,b)$ have different types, then no matter what the task type is, at least one worker's response is equal to a random coin flip. Therefore the probability that $M_{ia} = M_{ib}$ is equal to $\tfrac12$ for any task $i$, such that $\Ind(M_{ia} = M_{ib})$ is a Bernoulli random variable with parameter $\tfrac12$. For a threshold of $\xi$, the probability that the estimated clusters exactly recovers the $d$ worker types is at least as large as the probability that for all worker pairs $(a,b)$ of the same type, $\tfrac{1}{R} \sum_{i \in S} \Ind(M_{ia} = M_{ib}) > \xi$, and for all workers pairs $(a,b)$ of different types, $\tfrac{1}{R} \sum_{i \in \cS} \Ind(M_{ia} = M_{ib}) \leq \xi$. This can be loosely bounded by using Chernoff's bound for concentration of the average of Bernoulli random variables in addition to the union bound over all worker pairs. 
\begin{align*}
&1 - \textstyle\sum_{a \neq b} \Prob\Big(\tfrac{1}{R} \textstyle\sum_{i \in \cS} \Ind(M_{ia} = M_{ib}) \leq \xi\Big) \Ind(w_a = w_b) \\
&\qquad- \textstyle\sum_{a \neq b} \Prob\Big(\tfrac{1}{R} \textstyle\sum_{i \in \cS} \Ind(M_{ia} = M_{ib}) > \xi\Big) \Ind(w_a \neq w_b) \\
&= 1 - \exp\Big(- 2(\tfrac12 + \tfrac{(2 p - 1)^2}{2d} - \xi)^2 R\Big) \textstyle\sum_{a \neq b} \Ind(w_a = w_b) \\
&\qquad- \exp\Big(- 2(\tfrac12 - \xi)^2 R\Big)\textstyle\sum_{a \neq b}  \Ind(w_a \neq w_b) .
\end{align*}
If we choose $\xi = \tfrac12 + \tfrac{( 2 p - 1)^2}{4 d}$, then the probability of exactly recovering the partition of worker types is bounded below by
\begin{align*}
&1 - \tbinom{W}{2} \exp\left(-\tfrac{R ( 2 p - 1)^4}{8 d^2} \right) .
\end{align*}

We also need to guarantee that the number of workers per type is at least equal to $L$ so that it is possible to find at least $L$ workers per type. The number of workers of type $z$ is given by $|\cW_z| = \sum_{j \in [W]} \Ind(w_j = z)$, which is a Binomial$(W,\tfrac{1}{d})$ random variable. By Chernoff's bound and union bound,
\begin{align*}
\Prob(\cup_{z \in [d]} \{|\cW_z| < L\}) 
&\leq \sum_{z\in[d]} \Prob(|\cW_z| \leq L) \\
&\leq d \exp(-2 (1 - \tfrac{Ld}{W})^2 \tfrac{W}{d}).
\end{align*}

In order to analyze the estimates of the task solution, we show that the distribution of responses collected within each worker type for each task concentrates to its expectation. Let $S_{iz}$ be defined as
\begin{align*}
S_{iz} &:= \textstyle\sum_{j \in \cN(i) \cap \cW_z} \Ind(M_{ij} = +1) \\
&= \textstyle\sum_{j \in \cN(i) \cap \cW_z} \tfrac{1+ M_{ij} }{2} \\
&= L\left(\tfrac12 + \tfrac{1}{L}\textstyle\sum_{j \in \cN(i) \cap \cW_z} M_{ij}\right),
\end{align*}
such that $\sum_{j \in \cN(i) \cap \cW_z} M_{ij} = L (\tfrac{S_{iz}}{L} - \tfrac{1}{2})$. By model assumptions, the distribution of $S_{iz}$ will be Binomial($|\cN(i) \cap \cW_z|,\tfrac12$) if $t_i \neq z$, i.e. the worker type is different than the task type. If $t_i = z$, i.e. the worker type is equal to the task type, then the distribution of $S_{iz}$ will either be Binomial($|\cN(i) \cap \cW_z|,p$) if the task solution $a_i$ is equal to $+1$ and Binomial($L,1-p$) if the task solution $a_i$ is equal to $-1$. We bound the probability that $S_{iz}$ concentrates around its expectation within an additive error of $\tfrac12 |p - \tfrac12||\cN(i) \cap \cW_z|$ using Chernoff's bound,
\begin{align*}
&\Prob\left(|S_{iz} - \E[S_{iz}]| \geq \tfrac12 |p - \tfrac12||\cN(i) \cap \cW_z|\right) \\
&\leq 2 \exp\Big(-2 \Big(\tfrac12 |p - \tfrac12|\Big)^2 |\cN(i) \cap \cW_z|\Big) \\
&= 2 \exp\left(- \tfrac{(2 p - 1)^2 |\cN(i) \cap \cW_z|}{8} \right).
\end{align*}

Conditioned on the events that the worker cluster partitioning is exactly correct, i.e. $C = d$ and $\cV_z = \cW_z$ for all $z \in [d]$, and that $S_{iz}$ concentrates within an additive error of $\tfrac12 |p - \tfrac12|$ around its expectation for all $z \in [d]$, it follows that $z^*(i) = t_i$ and $a_i = \sgn\left(\sum_{j \in \cN(i) \cap \cW_{t_i}} M_{ij}\right)$. Recall that by construction $|\cN(i) \cap \cW_z| = L$ for all $i \in [T] \setminus \cS$ and $z \in [d]$, and  $|\cN(i) \cap \cW_z| = |\cW_z|$ for $i \in \cS$. Conditioned on the event that $|\cW_z| \geq L$ for all $z \in [d]$, this implies that $|\cN(i) \cap \cW_z| \geq L$ for all $i \in [T] \setminus \cS$ and $z \in [d]$.
The expected fraction of incorrect responses is bounded by
\begin{align*}
&\E\left[\tfrac{1}{T}\textstyle\sum_{i \in [T]} \Ind(\hat{a}_i \neq a_i)\right]  \\
&\leq \tbinom{W}{2} \exp\left(-\tfrac{R ( 2 p - 1)^4}{8 d^2} \right) + d \exp(-2 (1 - \tfrac{Ld}{W})^2 \tfrac{W}{d}) \\
&\quad + 2 d \exp\left(- \tfrac{(2 p - 1)^2 L}{8} \right).
\end{align*}

To limit the fraction of errors to $\alpha$, we can choose $R = \tfrac{8 d^2}{ ( 2 p - 1)^4} \ln\left(\tfrac{3 W(W-1)}{2 \alpha}\right), L = \tfrac{8}{(2p - 1)^2} \ln\left(\tfrac{6 d}{\alpha}\right),$
and
$W > \tfrac{16d}{(2p - 1)^2} \ln\left(\tfrac{6 d}{\alpha}\right)$.
The average number of solicited responses per task average out to 
\begin{align*}
&\tfrac{1}{T}(WR + Ld(T-R)) 
\leq Ld + \tfrac{WR}{T} \\ 
&= \tfrac{8d}{(2p - 1)^2} \ln\left(\tfrac{6d}{\alpha}\right) + \tfrac{8 W d^2}{ T ( 2 p - 1)^4} \ln\left(\tfrac{3W(W-1)}{2\alpha}\right).
\end{align*}
With a sufficiently large constant $C'$ such that $T \geq C' W^3$, the second term is dominated by the first term in the above equations, showing the desired result.

\section{Result II: Graphon Lower Bound}

In this section, we illustrate that we can use the reduction between crowdsourcing and graphon estimation in order to produce lower bounds for graphon estimation, since there are known lower bounds in the crowdsourcing literature. Let us consider the ``spammer-hammer'' model, i.e. $F(i,j) = w_j \in \{\tfrac12, 1\}$ for all $i,j$, and let $\sigma^2 := \tfrac{1}{W} \sum_{j \in [W]} (2 w_j - 1)^2 = \tfrac{1}{W} \sum_{j \in [W]} \Ind(w_j = 1)$. This model assumes that each collected response is either completely useless (equivalent to a coin toss), or it is 100 percent reliable and true. The challenge now becomes to filter out and distinguish the high quality responses from the noise. A minimax lower bound for nonadaptive algorithms for the Dawid-Skene model has been shown using the spammer-hammer model, since a task can only be answered correctly if there was at least one response collected from a ``hammer'' \cite{KargerOhShah14}. We recap the argument presented in their paper. Since the data is collected without knowledge of who is a spammer vs. hammer, the probability of collecting responses from all spammers is at least $(1-\sigma^2)^{pW}$, where $pW$ is the number of responses collected. Thus for any estimator (assuming $\sigma^2 \leq \tfrac{2}{3}$), 
\begin{align*}
\tfrac{1}{T} \textstyle\sum_{i \in [T]} \Prob(\hat{a}_i \neq a_i) \geq \tfrac{1}{2}(1-\sigma^2)^{pW} \geq \tfrac{1}{2} e^{-(\sigma^2 + \sigma^4)pW}
\end{align*}
Specifically, for a task $i$ assigned to $L$ workers,
\[\Prob(\hat{a}_i \neq a_i) \geq \tfrac{1}{2}(1-\sigma^2)^{L} \geq \tfrac{1}{2} e^{-(\sigma^2 + \sigma^4)L}.\]

If we use the estimator described \eqref{eq:max_est}, which estimates $\hat{M} \approx \E[M]$, and chooses $\hat{a}_i$ based on the response of the worker which maximizes $|\hat{M}_{ij}|$, then
\begin{align*}
 \Prob\left( \|\hat{M} - M\|_{\infty} \geq \tfrac{1}{2} \right) &\geq \Prob(\hat{a}_i \neq a_i) 
\geq \tfrac{1}{2} e^{-(\sigma^2 + \sigma^4)pW}.
\end{align*}

This also implies a lower bound on the MSE,
\begin{align}
\MSE 
&= \tfrac{1}{T} \textstyle\sum_{i \in [T]} \E\left[\tfrac{1}{W} \textstyle\sum_{j \in [W]} (\hat{M}_{ij} - M_{ij})^2\right] \nonumber \\
&\geq \tfrac{1}{T} \textstyle\sum_{i \in [T]} \Prob\left( \bigcup_{j \in [W]} \left\{|\hat{M}_{ij} - M_{ij}| \geq \tfrac{1}{2} \right\} \right) \tfrac{1}{4W} \nonumber \\
&\geq  \tfrac{1}{8W} \exp\left(-(\sigma^2 + \sigma^4)pW \right). \label{eq:DS_LB}
\end{align}

\medskip
\noindent{\bf Construct an Equivalent Graphon Estimation Task.}
Given an instance of the crowdsourcing task under the spammer-hammer model, we construct a graphon estimation task such that the graphon data matrix is statistically equivalent to the collected crowdsourced response data matrix. If there existed an algorithm for graphon estimation that could achieve a given precision, it would directly imply that one could estimate the corresponding expected crowdsourced response matrix within the same precision given a uniformly random assignment scheme. The lower bound of \eqref{eq:DS_LB} would then imply a lower bound for the constructed graphon estimation task.

Consider a spammer-hammer model as described previously, where $F_{ij} = w_j \in \{0,1\}$, and we assume a worker $j$ is a spammer ($w_j = \tfrac12$) with probability $1 - \sigma^2$ and a hammer $(w_j = 1)$ with probability $\sigma^2$. Let $\alpha$ be a prior on the task answer, such that $a_i = +1$ with probability $\alpha$ and $a_i = -1$ with probability $1 - \alpha$. Assume that the number of workers is approximately proportional to the number of tasks, such that $T \sim$ Binomial$(n, \beta)$ and $W = n - T$ for some $n$. Note that $T + W = n$ and $\E[T/(T+W)] = \beta$. As $n \to \infty$, the proportion of workers to tasks converges, i.e. $T \to \beta n$ and $W \to (1 - \beta) n$. 

Since the collected crowdsourced response matrix is asymmetric, we consider a larger symmetric $(T+W)\times(T+W)$ matrix where the upper right $T \times W$ block of the matrix is $M$ and the lower right $W \times T$ block of the matrix is $M^T$. We consider the graphon estimation task with centered binary data matrices which take values $\{-1,+1\}$ rather than $\{0,1\}$, which simplifies the notation in translating between the crowdsourcing model and the matrix estimation model. The (centered) graphon model assumes that each $i \in [T+W]$ is associated to a latent variable $\theta_i$ sampled uniformly on the unit interval. For uniformly sampled entries indexed by $\cE \in [(T+W) \times (T+W)]$, the observed data distribution when centered is 
\begin{align*}
Y_{ij} = 
\begin{cases}
+1 &\mbox{ with probability } \tfrac12(1 + f(\theta_i,\theta_j)) \\
-1 &\mbox{ with probability } \tfrac12(1 - f(\theta_i,\theta_j)).
\end{cases}
\end{align*}
such that $\E[Y_{ij}] = f(\theta_i,\theta_j)$. 

Recall for the spammer hammer model that $\E[M_{ij}] = a_i (2 w_j - 1)$ takes value 0 if $w_j = \tfrac12$, and otherwise takes value $a_i$. We represent our worker or task types with the latent parameter $\theta_i \in [0,1]$. Split $[0,1]$ into four intervals, specified by $\cI_1 = [0, \alpha\beta], \cI_2 = [\alpha\beta, \beta], \cI_3 = [\beta, 1 - \sigma^2 (1 -\beta)],$ and $\cI_4 = [1 - \sigma^2 (1 -\beta), 1]$. $\cI_1$ corresponds to a task with +1 answer, $\cI_2$ corresponds to a task with $-1$ answer, $\cI_3$ corresponds to a spammer worker, and $\cI_4$ corresponds to a hammer worker. Construct the latent function $f$ according to
\begin{align*}
f(\theta_i, \theta_j)
&=\begin{cases}
\sgn(\alpha\beta - \theta_i) &\mbox{if } \theta_i \in \cI_1 \cup \cI_2, \theta_j \in \cI_4 \\
\sgn(\alpha\beta - \theta_j) &\mbox{if } \theta_i \in \cI_4, \theta_j \in \cI_1 \cup \cI_2 \\
0 &\mbox{otherwise }.
\end{cases}
\end{align*}

We can verify that the data matrix $Y$ randomly generated from latent variables $\theta_i$ and constructed latent function $f$ is statistically equivalent (under permutation) to the block matrix $\left[\begin{smallmatrix}Q &M\\M^T& Q'\end{smallmatrix}\right]$, where $M$ is the collected response matrix from the corresponding spammer-hammer crowdsourced model, and $Q, Q'$ are $T \times T$ and $W \times W$ binary symmetric matrices respectively with observed entries taking value +1 or -1 with equal probability. Thus an estimate of $\E[Y]$ directly translates to an estimate of $\E[M]$, such that a graphon estimation algorithm that achieves some precision for this latent function $f$ implies that there exists an algorithm for estimating the expected crowdsourced response matrix $\E[M]$ for the corresponding spammer-hammer model with parameter $\sigma^2$. 

For the symmetric function $f: [0,1]^2 \to \Reals$, one can consider the corresponding Hilbert–Schmidt integral operator $F: L^2([0,1],\Reals) \to L^2([0,1],\Reals)$ which operates over $L^2$ functions over the unit interval. Since the operator is compact and self-adjoint, the operator $F$ has a discrete spectrum such that 
\[f(\theta_i, \theta_j) = \textstyle\sum_{k \in \IntegersP} \lambda_k q_k(\theta_i) q_k(\theta_j), \]
for orthonormal eigenfunctions $q_k$ and eigenvalues $\lambda_k$, i.e. $\langle q_k, q_{k'} \rangle = 0$ for $k \neq k'$ and $\langle q_k, q_k \rangle = 1$ for all $k$, where
\[\langle q_k, q_{k'} \rangle = \int_0^1 q_k(\theta) q_{k'}(\theta) d\theta.\]
In the next two sections, we provide minimax lower bounds for graphon estimation with respect to related spectral properties of $f$. 

\medskip
\noindent{\bf Lower bounds with respect to maximum eigenfunction magnitude.}
Let $\cF_B$ indicate the class of graphon models, represented by the associated set of latent functions $f$, for which $\sup_{k \in \IntegersP} \sup_{\theta \in [0,1]} |q_k(\theta)| = B$, where $\{q_k\}_{k \in \IntegersP}$ are the eigenfunctions of the integral operator associated to $f$. We compute the minimax bounds on the mean squared error and the probability that the max error is larger than one half. For a fixed $p$ (density of observations), $n$, and $B \in [1,\infty)$, we minimize the error over all choices of estimators $\hat{Y}$, which is a function that takes in a data matrix $Y$ and produces an estimated matrix $\hat{Y} \approx \E[Y]$, and we maximize the error over all models in $\cF_B$, where the models are specified by $f(\theta_i, \theta_j)$. The minimax lower bounds comes from constructing an instance of the spammer-hammer crowdsourcing model such that $\sup_{\theta \in [0,1]} |q_k(\theta)| = B$, and then applying the above lower bounds from the crowdsourcing setting.

\begin{theorem} \label{thm:LB_B}
Let $\cF_B$ be the class of latent variable models where the supremum of the eigenfunction amplitudes is $B \in [1,\infty)$. Let $Y$ be the $n \times n$ data matrix generated according to the centered graphon latent variable model, and let $p$ be the density of entries observed in the matrix uniformly at random. Let $\hat{Y}$ be an estimator which takes in the partially observed data matrix $Y$ and produces an estimated matrix $\hat{Y}$.
\begin{align*}
\min_{\hat{Y}} \max_{\cF_B} \Prob\left(\|\hat{Y} - \E[Y]\|_{\infty} \leq \tfrac12\right) \geq \tfrac{1}{2} \exp\left(-\tfrac{pn}{2B^2 - 1}\right), \\
\min_{\hat{Y}} \max_{\cF_B} \E\left[\tfrac{1}{n^2} \textstyle\sum_{ij} (\hat{Y}_{ij} - \E[Y])^2 \right] 
\geq \frac{B^2 e^{-\tfrac{pn}{2B^2 -1}}}{4(2B^2 - 1) n}.
\end{align*}
\end{theorem}

These lower bounds follow from choosing $\beta = \tfrac{1}{2B^2}$ and $\sigma^2 = \tfrac{1}{2B^2 - 1}$ for the spammer-hammer model. Previous lower bounds have looked specifically at the scaling of $n$ or $pn$ with respect to the rank of the matrix, but this lower bound focuses on showing the scaling with respect to $B$, or the maximum amplitude of the eigenfunctions. This result implies that $pn$ must scale as $B^2$. The property $B$ is related to the incoherence property as well.

\medskip
\noindent{\bf Lower bounds with respect to minimum eigenvalue.}
We can also show a lower bound in relation to the eigenvalues much in the same way. Let $\cF_{\lambda}$ indicate the set of latent variable models for which $\sup_{k \in \IntegersP} |\lambda_k| = \lambda$. We compute the minimax bounds on the mean squared error and the probability that the max error is larger than one half. For a fixed $p$, $n$, and $\lambda \in [0,\tfrac{1}{2}]$, we minimize the error over all choices of estimators $\hat{Y}$, which is a function that takes in a data matrix $Y$ and produces an estimated matrix $\hat{Y}$, and we maximize the error over all models in $\cF_{\lambda}$, where the models are specified by $f(\theta_i, \theta_j)$. The minimax lower bounds comes from constructing an instance of the spammer-hammer crowdsourcing model such that  $\sup_{k \in \IntegersP} |\lambda_k| = \lambda$, and then applying the above lower bounds from the crowdsourcing setting.

\begin{theorem} \label{thm:LB_lambda}
Let $\cF_{\lambda}$ be the class of latent variable models where the minimum magnitude nonzero eigenvalue has magnitude $\lambda \in [0,\tfrac12]$. Let $Y$ be the $n \times n$ data matrix generated according to the latent variable model, and where $p$ is the density of entries observed in the matrix. Let $\hat{Y}$ be an estimator which takes in the partially observed data matrix $Y$ and produces and estimated matrix $\hat{Y}$.
\begin{align*}
\min_{\hat{Y}} \max_{\cF_{\lambda}} \Prob\left(\|\hat{Y} - \E[Y]\|_{\infty} \leq \tfrac12\right) 
\geq  \tfrac{1}{2} e^{-2 \lambda^2 (4 \lambda^2 + 1) pn}, \\
\min_{\hat{Y}} \max_{\cF_{\lambda}} \E\left[\tfrac{1}{n^2} \textstyle\sum_{ij} (\hat{Y}_{ij} - \E[Y])^2 \right] 
\geq \tfrac{1}{4n} e^{-2 \lambda^2 (4 \lambda^2 + 1) pn}.
\end{align*}
\end{theorem}

These lower bounds follow from choosing $\sigma^2 = 4 \lambda^2$ and $\beta = \tfrac12$ for the spammer-hammer model. This result implies that $pn$ must scale as $\lambda_{\min}^{-2}$.

\section{Conclusion}

In this paper, we showed that the crowdsourcing inference problem can be statistically reduced to graphon or $1$-bit matrix
estimation problem. This helps in multiple ways. First, it helps us show that it is indeed feasible to beat the majority voting
baseline performance for setup where the underlying model leads to rank $d$ setup for $d > 1$. This is in contrast to all
known theoretical results that have established such property as all known results have been about rank $1$ model. Second,
it helps us establish refined, explicit lower bounds for graphon estimation problem by invoking lower bound for crowdsourcing
problem. This allows us to obtain a useful invariant: the number of samples multiplied by the square of the minimum eigenvalue
associated with the spectral decomposition of the graphon model is always lower bounded by a universal constant. Going forward,
we believe that this relationship will help develop better estimation algorithms for the crowdsourcing problem. Further it will advance our understanding of graphon estimation as the crowdsourcing problem requires only row-sum estimation, not the entire matrix.

\subsubsection*{Acknowledgments}

This work is supported in parts by a Microsoft Research postdoctoral fellowship and NSF projects CMMI 1462158, CMMI 1634259, CNS 1523546 and ARO MURI W911NF-16-1-0551.  

\bibliographystyle{plain}
\bibliography{bibliography} 

\appendix

\section{Proving the Graphon Minimax Lower Bounds}

The first step of the proof is to study the spectral properties of the constructed graphon model as a function of the spammer-hammer crowdsourcing model.

\medskip \noindent
{\bf Spectral Properties of Graphon Function Constructed from Spammer-Hammer Model.}
Recall that for a spammer-hammer model with $n = W+T$, $\E[T] = \beta n$, $a_i = +1$ with probability $\alpha$, and $w_j \in \{\frac12, 1\}$ such that $w_j = 1$ with probability $\sigma^2$, we constructed the following graphon latent function
\begin{align*}
f(\theta_i, \theta_j)
&=\begin{cases}
\sgn(\alpha\beta - \theta_i) &\mbox{if } \theta_i \in \cI_1 \cup \cI_2, \theta_j \in \cI_4 \\
\sgn(\alpha\beta - \theta_j) &\mbox{if } \theta_i \in \cI_4, \theta_j \in \cI_1 \cup \cI_2 \\
0 &\mbox{otherwise },
\end{cases}
\end{align*}
where $\cI_1 = [0, \alpha\beta], \cI_2 = [\alpha\beta, \beta], \cI_3 = [\beta, 1 - \sigma^2 (1 -\beta)],$ and $\cI_4 = [1 - \sigma^2 (1 -\beta), 1]$. We show that the integral operator corresponding to $f$ has finite spectrum with rank $2$. 

It has eigenvalues $\lambda_1 = \sqrt{\sigma^2 \beta (1 - \beta)}$ and $\lambda_2 = -\sqrt{\sigma^2 \beta (1 - \beta)}$, and the corresponding eigenfunctions are
\begin{align*}
q_1(\theta) &= 
\begin{cases}
\tfrac{\sgn(\alpha\beta - \theta)}{\sqrt{2 \beta}}  &\mbox{if } \theta \in \cI_1 \cup \cI_2 \\
0  &\mbox{if } \theta \in \cI_3 \\
\tfrac{1}{\sqrt{2 \sigma^2 (1 - \beta)}} &\mbox{if } \theta \in \cI_4 
\end{cases} \\
q_2(\theta) &= 
\begin{cases}
\tfrac{\sgn(\alpha\beta - \theta)}{\sqrt{2 \beta}} &\mbox{if } \theta \in \cI_1 \cup \cI_2 \\
0  &\mbox{if } \theta \in \cI_3 \\
- \tfrac{1}{\sqrt{2 \sigma^2 (1 - \beta)}} &\mbox{if } \theta \in \cI_4 .
\end{cases}
\end{align*}
We can verify that $q_1$ and $q_2$ are orthonormal, and that they indeed form a spectral representation for $f$, 
\begin{align*}
&\int_0^1 q_1(\theta)^2 d\theta = \int_0^1 q_2(\theta)^2 d\theta  = 1 \\
&\int_0^1 q_1(\theta) q_2(\theta) d\theta = 0 \\
&f(\theta_i,\theta_j) = \lambda_1 q_1(\theta_i) q_1(\theta_j) + \lambda_2 q_2(\theta_i) q_2(\theta_j) .
\end{align*}
The supremum amplitude of the eigenfunctions is $B = \max_{\theta \in [0,1], k \in [2]} |q_k(\theta)| = (2 \min(\beta, \sigma^2 (1 - \beta)))^{-1/2}$.

\medskip \noindent
{\bf Proof of Theorem \ref{thm:LB_B}: Lower bounds with respect to maximum eigenfunction magnitude.}
In order to prove the minimax lower bounds, for each value of $B \in [1,\infty)$, we construct an instance of the spammer-hammer crowdsourcing model which lies within $\cF_B$, and then we apply the previous crowdsourcing lower bounds. Recall the characterization presented above for transforming the data matrix from the spammer-hammer model to the graphon model. For some $B \in [1,\infty)$, we choose $\sigma^2 = (2B^2 - 1)^{-1}$ and $\beta = (2B^2)^{-1}$, such that we can verify indeed $B = (2 \min(\beta, \sigma^2 (1 - \beta)))^{-1/2}$. The lower bound on the error probability follows from plugging in these quantities to the corresponding previous lower bound
\begin{align*}
&\tfrac{1}{2} \exp\left(-(\sigma^2 + \sigma^4)pW\right) \\
&= \tfrac{1}{2} \exp\left(-\left(\tfrac{1}{2B^2 - 1} + \tfrac{1}{(2B^2 - 1)^2}\right) p (1-\beta) n\right) \\
&= \tfrac{1}{2} \exp\left(- \tfrac{2B^2}{(2B^2 - 1)^2} p \tfrac{2B^2 - 1}{2B^2} n\right) \\
&= \tfrac{1}{2} \exp\left(- \tfrac{p n}{(2B^2 - 1)}\right).
\end{align*}
The lower bound for the MSE similarly follows from plugging in these quantities to the corresponding previous lower bound
\begin{align*}
\tfrac{1}{8W} e^{-(\sigma^2 + \sigma^4)pW}
&= \tfrac{1}{8(1-\beta)n} \exp\left(- \tfrac{p n}{(2B^2 - 1)}\right) \\
&= \tfrac{B^2}{4 (2 B^2 - 1) n} \exp\left(- \tfrac{p n}{(2B^2 - 1)}\right).
\end{align*}

\medskip \noindent
{\bf Proof of Theorem \ref{thm:LB_lambda}: Lower bounds with respect to minimum eigenvalue.}
In order to prove the minimax lower bounds, for each value of $\lambda \in [0,\tfrac12]$, we construct an instance of the spammer-hammer crowdsourcing model which lies within $\cF_{\lambda}$, and then we apply the previous lower bounds. Recall the characterization presented above for transforming the data matrix from the spammer-hammer model to the graphon model. For some $\lambda \in [0,\tfrac12]$, we choose $\sigma^2 = 4 \lambda^2$ and $\beta = \tfrac12$, such that we can verify indeed $\min_k |\lambda_k| = \lambda$. The lower bound on the error probability follows from plugging in these quantities to the corresponding previous lower bound
\begin{align*}
\tfrac{1}{2} \exp\left(-(\sigma^2 + \sigma^4)pW\right)
&= \tfrac{1}{2} \exp\left(-(4 \lambda^2 + 16 \lambda^4) p \tfrac{1}{2} n\right) \\
&= \tfrac{1}{2} \exp\left(- 2 \lambda^2 ( 4 \lambda^2 + 1) pn\right).
\end{align*}
The lower bound for the MSE similarly follows from plugging in these quantities to the corresponding previous lower bound
\begin{align*}
\tfrac{1}{8W} e^{-(\sigma^2 + \sigma^4)pW}
&= \tfrac{1}{4 n} \exp\left(- 2 \lambda^2 ( 4 \lambda^2 + 1) pn\right).
\end{align*}

\end{document}